\documentclass{article}
\makeatletter
\def\input@path{{other/}}
\makeatother
\usepackage{other/fancyhdr}
\usepackage{other/natbib}
\makeatletter
\@namedef{ver@fancyhdr.sty}{}
\@namedef{ver@natbib.sty}{}
\makeatother
\usepackage{iclr2027_conference,times}
\usepackage[T1]{fontenc}
\usepackage{amsmath,amssymb,graphicx,booktabs,multirow,makecell}
\usepackage{float}
\usepackage{xcolor}
\usepackage[hidelinks]{hyperref} 
\usepackage{url}
\usepackage{wrapfig,booktabs,makecell}

\title{Dynamic Merging Mutually-Distilled Multi-View Networks for RGB-Event-based Representation Learning}

\author{Futian Wang$^{1}$, Fan Zhang$^{1}$, Xiao Wang$^{1}$~\thanks{Corresponding author: Xiao Wang, Yuhe Ding}, 
Yuhe Ding$^{2}$\footnotemark[1], Shiao Wang$^{1}$, Jingxi Huang$^{1}$, Jin Tang$^{1}$ \\ 
1. School of Computer Science and Technology, Anhui University, Hefei 230601, China \\ 
2. Hefei University, Hefei 230601, China \\ 
\texttt{\{wft, xiaowang, tangjin\}@ahu.edu.cn} \\  
\texttt{e24301169@stu.ahu.edu.cn} \\
\texttt{madao3c@foxmail.com} \\
\texttt{e24101001@stu.ahu.edu.cn} \\
\texttt{e22314020@stu.ahu.edu.cn}
}

\iclrfinalcopy 
\begin{document}

\maketitle
\lhead{\small Dynamic Merging Mutually-Distilled Multi-View Networks for RGB-Event-based Representation Learning}

\begin{abstract}
Multi-modal fusion algorithms combining conventional frame-based cameras and event cameras have recently attracted substantial research attention. Typically, existing approaches convert event streams into frames, voxels, or tensors, and then learn the required spatiotemporal information from these transformed event representations for fusion with conventional frame-camera data. Obviously, such paradigms only exploit the limited representational capacity of event data. Moreover, standard training pipelines fail to fully leverage diverse model parameters obtained throughout the learning process. In this paper, we propose a novel mutual distillation method leveraging multi-view teacher network weights to better integrate multiple event representations and achieve model-parameter level fusion. Specifically, we employ multiple encoders to extract different types of event representations and perform mutual knowledge distillation at both the feature-level and logit-level in this process. Afterwards, we perform a dynamic hierarchical model merge strategy on the model parameters from the first stage to yield a unified multi-modal RGB-Event representation model. Experimental results across multiple event-based datasets demonstrate that our proposed framework achieves significant performance improvements.
The source code of this paper will be released at \url{https://github.com/Event-AHU/Multi-View_Distilled_Merge}.
\end{abstract}

\section{Introduction}

RGB video offers rich appearance and scene context through regularly sampled frames, forming the basis of many visual recognition systems. However, its fixed temporal sampling can miss rapid motion and fine temporal changes. Event cameras instead record asynchronous brightness changes with high temporal resolution and wide dynamic range, offering complementary motion cues. Combining the two sensing modalities therefore provides a way to associate detailed appearance with temporally precise dynamics.

\begin{figure}[!t]
\centering
\includegraphics[width=1\linewidth]{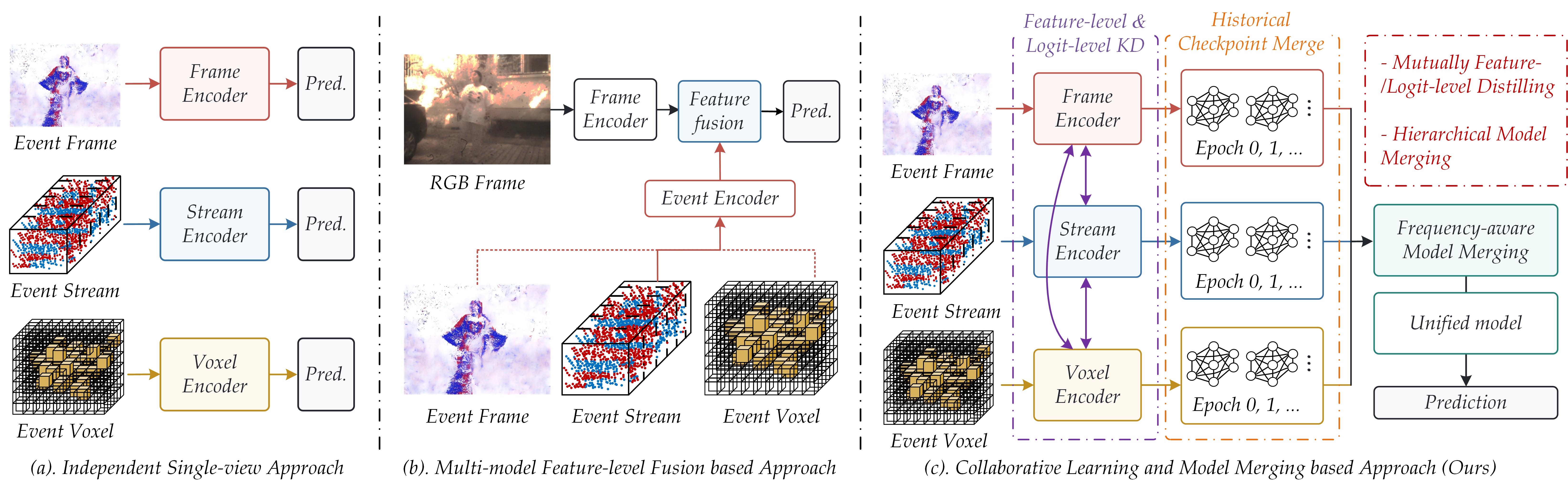}
\caption{Comparison of RGB-Event representation-learning paradigms. (a) RGB-Event fusion with one event representation. (b) Our multi-view framework combines mutual distillation across event-frame, event-stream, and event-voxel branches with dynamic hierarchical model merging.}  
\label{fig:fig1}
\end{figure}

Event observations can be represented as accumulated event frames, asynchronous event streams, or spatio-temporal voxel grids. Event frames preserve spatial response patterns, streams retain precise timing, and voxels organize local spatial and temporal information. These representations have different structures and statistics, thus, selecting only one view or fusing views without accounting for their differences can discard useful event information. How can we learn a coherent RGB-Event representation while preserving the complementary cues carried by multiple event views?

Learning from multiple event views requires reconciling shared semantics with representation-specific characteristics. Their heterogeneous structures induce different feature distributions, so indiscriminate alignment may weaken the diversity that motivates multi-view learning. Cross-modal distillation offers a means of transferring supervision between modalities~\citep{wang2021evdistill,huo2024c2kd}, while mutual learning allows peer networks to learn collaboratively without a fixed teacher~\citep{zhang2018deep}. Together, these perspectives motivate knowledge exchange that promotes semantic agreement while respecting the distinct information conveyed by each event view.

Complementarity also extends to the parameters acquired during learning. Models associated with different event views and different stages of optimization can encode distinct aspects of the training observations. Model merging provides a basis for integrating such knowledge through parameter updates~\citep{ilharco2022editing}, with subsequent work addressing interference and the separation of shared and specialized components~\citep{yadav2023ties,lu2024twin,zheng2025free}. Nevertheless, a common backbone architecture does not ensure that updates learned from heterogeneous event inputs are mutually compatible. The central challenge is to integrate knowledge across event representations and optimization histories without erasing the specialization that makes these sources complementary.

Motivated by this perspective, we propose a mutually distilled multi-view framework for RGB-Event representation learning. As illustrated in Fig.~\ref{fig:fig1}, the framework extends single-view learning and conventional cross-modal fusion by considering complementarity in both representation space and parameter space. Collaborative learning allows event-frame, event-stream, and event-voxel branches to exchange knowledge through feature- and logit-level mutual distillation. Dynamic hierarchical merging then integrates historical optimization knowledge with frequency-aware cross-view parameter consolidation. This formulation connects the acquisition of complementary representations with the preservation of shared and view-specific knowledge in a unified model.

To sum up, the main contributions of this paper can be summarized as follows: 

1). We propose a novel collaborative multi-view event data representation (i.e., event voxels, frames, and streams) learning framework to tackle monotonous event representations, weak cross-modal feature coupling and inadequate cross-modal information mining. Both feature- and score-level bidirectional mutual distillation strategies are adopted to facilitate mutual supervision and collaborative optimization among the three fusion branches. 

2). We propose a dynamic hierarchical model merge strategy for multi-modal and multi-view RGB-Event fusion. This strategy fully exploits model weights from different training epochs and performs frequency-aware model merging using the optimal weights of the three-branch model.  

3). Extensive experiments on two widely used RGB-Event based recognition benchmark datasets (i.e., PokerEvent~\citep{wang2025sstformer}, HARDVS~\citep{wang2024hardvs}) fully validated the effectiveness of our proposed dynamic hierarchical model merge strategy for this task.


\section{Related Works} \label{sec::relatedWorks}

\textbf{Event-based Vision.~} 
Event cameras record asynchronous brightness changes with high temporal resolution and wide dynamic range, motivating RGB-Event methods that combine appearance with precise motion cues. Recent methods include hybrid spiking-transformer processing in SSTFormer~\citep{wang2025sstformer}, CNN--Transformer interaction in TSCFormer~\citep{wang2025unleashing}, multimodal heat-conduction fusion in MMHCO-HAR~\citep{wang2025human}, and hardware-oriented event processing in ViP-HMNN~\citep{ji2025vip}. Most existing methods, however, use a single fixed event representation, such as frames or voxels. Because event frames, streams, and voxels encode different spatial and temporal information, our method uses parallel view-specific branches and mutual distillation to preserve and integrate their complementarity.

\textbf{Model Merging.~} 
Model merging combines fine-tuned networks into one model without joint retraining. Task Arithmetic~\citep{ilharco2022editing} introduced task vectors and weighted parameter addition, while later methods addressed interference through parameter competition~\citep{du2024parameter}, global-basis projection~\citep{liu2026globa}, isotropic subspace alignment~\citep{marczak2025no}, learnable singular-component selection~\citep{lee2025adarank}, separation of shared and exclusive knowledge~\citep{lu2024twin}, and frequency-domain filtering~\citep{zheng2025free}. These methods generally target homogeneous model instances. Our approach integrates branches with a shared backbone and heterogeneous event inputs, combining parameter consolidation with mutual distillation for RGB-Event representation learning.

\textbf{Mutual Distillation.~} 
Knowledge distillation transfers supervision by aligning predictions or intermediate representations. Cross-modal approaches include bidirectional image-to-event distillation~\citep{wang2021evdistill}, object-centric feature distillation~\citep{li2024object}, dynamic bidirectional distillation for modality imbalance~\citep{huo2024c2kd}, and frequency-decoupled feature transfer~\citep{liu2024frequency}. Deep Mutual Learning~\citep{zhang2018deep} and subsequent online and competitive variants~\citep{guo2020online,shi2025competitive} allow peer networks to learn without a fixed teacher. Most such methods use logits or fixed peer aggregation. Our method combines logit-level supervision with sample-adaptive feature distillation, using peer confidence and cross-view similarity to weight the contribution of each event view.

\section{Methodology} \label{sec::method}

\subsection{Overview} 
\begin{figure}[!t]
\centering
\includegraphics[width=1\linewidth]{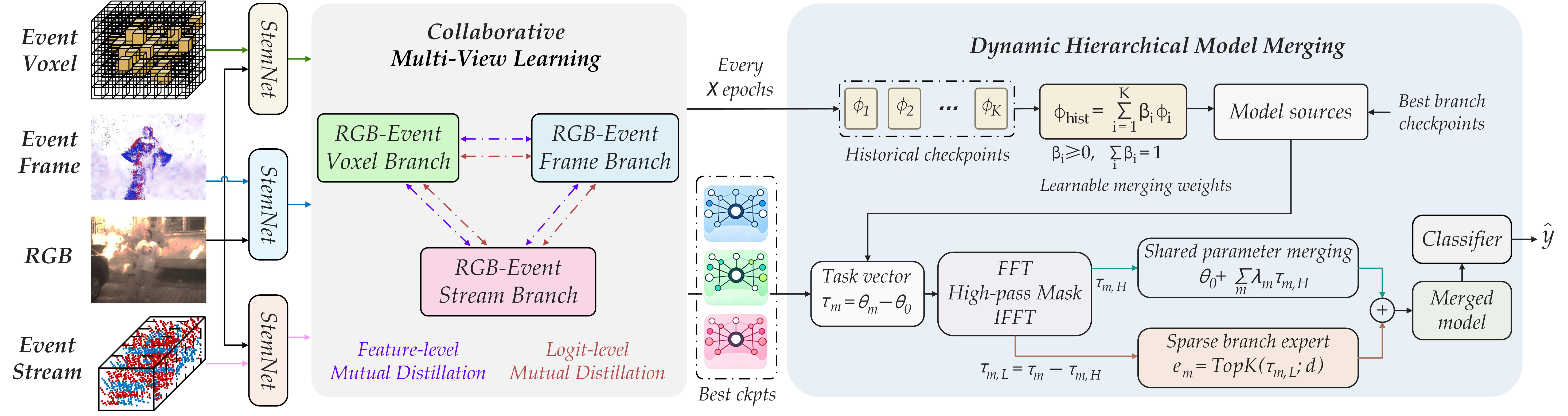}
\caption{Overview of the proposed RGB-Event representation learning framework. Three RGB-Event branches encode event frames, streams, and voxels and exchange knowledge through feature- and logit-level mutual distillation. Historical checkpoint aggregation and frequency-aware parameter merging then produce a unified inference model.}  
\label{fig:myframework}
\end{figure}

As shown in Fig.~\ref{fig:myframework}, our framework consists of three sequential components: Multi-View Event Encoding, Collaborative Multi-View Learning, and Dynamic Hierarchical Model Merging. 
First, the Multi-View Event Encoding module converts the raw event sequence into three temporally aligned representations. These complementary representations capture the spatial, asynchronous, and spatio-temporal characteristics of the same event sequence, and are projected into a unified feature space through view-specific encoding networks. 
Second, the Collaborative Multi-View Learning module consists of three parallel RGB-Event branches, each taking the RGB frames together with one event representation as input. The three branches are jointly optimized through feature-level and logit-level mutual distillation, enabling complementary knowledge to be exchanged across different event views while preserving their view-specific representations. 
Finally, the Dynamic Hierarchical Model Merging module hierarchically integrates the knowledge learned by different branches. Specifically, historical checkpoints are first dynamically aggregated through trainable weight averaging to obtain robust branch models, after which the resulting models are merged in the frequency domain to produce a unified model that preserves both shared knowledge and branch-specific expertise.

\subsection{Multi-View Event Encoding}

Given a paired RGB-Event video sample, we denote the sampled RGB frames as $\mathbf{R}=\{R_1,\ldots,R_T\}\in\mathbb{R}^{T\times 3\times H\times W}$, where $T$, $H$, and $W$ are the temporal length, frame height, and frame width. The event camera records asynchronous events aligned with the RGB frames, where each event is represented as $(x_i,y_i,t_i,p_i)$, and $(x_i,y_i)$, $t_i$, and $p_i$ denote the spatial coordinate, timestamp, and polarity.
The event frame view accumulates events within each sampled RGB interval into a dense frame-like representation:
\begin{equation}
\mathbf{E}_{f}\in\mathbb{R}^{T\times C_f\times H\times W},
\label{eq:event_frame_view}
\end{equation}
where $C_f=3$. This view preserves the spatial layout of event responses and can be processed with image-style spatial transformations.
The event stream view keeps the asynchronous event form and partitions events according to the sampled RGB intervals:
\begin{equation}
\begin{aligned}
\mathbf{E}_{s} &= \{\mathcal{S}_t\}_{t=1}^{T},  ~~~\mathcal{S}_t &= \{(x_i,y_i,\tau_i,p_i)\}_{i=1}^{N_t},
\end{aligned}
\label{eq:event_stream_view}
\end{equation}
where $\tau_i$ denotes the timestamp normalized within the $t$-th interval, and $N_t$ is the number of events in this interval. In implementation, the events in each interval are discretized into 12 temporal bins.
The event voxel view voxelizes events in the spatial-temporal space. We treat the normalized timestamp as the temporal coordinate and represent each RGB interval as a set of occupied voxel tokens:
\begin{equation}
\begin{aligned}
\mathbf{E}_{v} &= \{\mathcal{V}_t\}_{t=1}^{T},~~~
\mathcal{V}_t &= \{[V_x,V_y,V_\tau,\mathbf{V}_f]\}\in\mathbb{R}^{Q_t\times L},
\end{aligned}
\label{eq:event_voxel_view}
\end{equation}
where $Q_t$ is the number of occupied voxels in the $t$-th interval, $(V_x,V_y,V_\tau)$ denotes the voxel center coordinate in the $x$-$y$-$\tau$ space, and $\mathbf{V}_f$ is the voxel feature aggregated from events falling into the voxel. Each voxel token has $L=19$ dimensions, consisting of three coordinate values and a 16-dimensional voxel feature. The three representations therefore describe different organizations of the event source rather than a single homogeneous event tensor.

Since the three event views exhibit heterogeneous data structures and feature dimensionalities, they cannot be directly processed by same backbones. 
To bridge this representation gap, we employ three view-specific stem networks to project the event frame, stream, and voxel into a unified feature space. 
Formally, the encoded representations are obtained as
\begin{equation}
\mathcal{E}{m}=\operatorname{SN}_{m}(E_{m}), \quad m\in\{f,s,v\},
\label{eq}
\end{equation}
where $\operatorname{SN}_{m}(\cdot)$ denotes the stem network corresponding to the $m$ event view, $\{f, s, v\}$ denote the event frame, stream and voxel views, respectively. 
The resulting representations, denoted as $\mathcal{E}_f$, $\mathcal{E}_s$, and $\mathcal{E}_v$, share a common embedding dimension while preserving the distinctive characteristics of their respective event views. 
These unified representations are subsequently fed into the collaborative multi-view learning module for joint representation learning.

\subsection{Collaborative Multi-View Learning}
The Collaborative Multi-View Learning module aims to jointly exploit the complementary information provided by heterogeneous event views while preserving their view-specific characteristics. 
Specifically, three parallel RGB-Event branches, denoted as $\Theta_f$, $\Theta_s$, and $\Theta_v$, take the RGB frames $\mathbf{R}$ together with the encoded event representations $\mathcal{E}_f$, $\mathcal{E}_s$, and $\mathcal{E}_v$ as input, respectively. 
Although the three branches share an identical backbone architecture, each branch is independently optimized to learn representations tailored to its corresponding event view. 
During training, complementary knowledge is exchanged across branches through both logit-level and feature-level mutual distillation.
For each view $m\in\{f,s,v\}$, the corresponding model outputs a high-level feature and a class logit:
\begin{equation}
\begin{aligned}
\mathbf{h}_m,\mathbf{z}_m
&=
N_{\Theta_m}(\mathbf{R},\mathcal{E}_m).
\end{aligned}
\label{eq:network_output}
\end{equation}
$\mathbf{h}_m$ is used for feature-level mutual distillation, and $\mathbf{z}_m$ is used for logit-level mutual distillation and final prediction. 
At the logit level, each model receives online peer supervision from the other two branches. 
For each branch $\Theta_m$, we define its peer set $\mathcal{P}_m$ as the collection of the remaining two branches corresponding to the other event views, i.e.,
$\mathcal{P}_m=\{\,n \mid n\in\{f,s,v\},\, n\neq m\,\}$.
With distillation temperature $T_d$, the peer soft target is:
\begin{equation}
\tilde{\mathbf{p}}_{\mathrm{peer},m}^{T_d}
=
\operatorname{Softmax}
\left(
\frac{1}{T_d|\mathcal{P}_m|}
\sum_{n\in\mathcal{P}_m}
\operatorname{sg}(\mathbf{z}_n)
\right),
\label{eq:peer_logit_target}
\end{equation}
where $\operatorname{sg}(\cdot)$ denotes stop-gradient. 
The peer soft target is used to supervise the prediction of branch $\Theta_m$ through knowledge distillation. 
The temperature-scaled prediction of the current branch is defined as
$\mathbf{p}_m^{T_d}=\operatorname{Softmax}\left(\frac{\mathbf{z}_m}{T_d}\right)$,
where the temperature scaling softens the predicted probability distribution to facilitate knowledge transfer. 
The logit-level mutual distillation loss for branch $\Theta_m$ is then formulated as
\begin{equation}
\mathcal{L}_{D}^{m}=T_d^{2}D_{\mathrm{KL}}\left(\tilde{\mathbf{p}}_{\mathrm{peer},m}^{T_d},\mathbf{p}_{m}^{T_d}\right),
\label{eq:logit_kd_loss}
\end{equation}
where $D_{\mathrm{KL}}(\cdot,\cdot)$ denotes the Kullback--Leibler divergence. The factor $T_d^{2}$ follows the standard temperature-scaled knowledge distillation formulation to preserve the gradient magnitude during optimization.

At the feature level, each branch feature is projected into a shared semantic space, and a sample-adaptive peer teacher is built from the remaining branches:
\begin{equation}
\mathbf{q}_m
=G_m(\mathbf{h}_m),\ 
\tilde{\mathbf{q}}_m=\sum_{n\in\mathcal{M},\,n\neq m} a_{m,n}\operatorname{sg}(\mathbf{q}_n),
\label{eq:feature_teacher}
\end{equation}
where $G_m(\cdot)$ is the feature projector of branch $m$. The adaptive weight $a_{m,n}$ is predicted from the student feature, peer feature, peer confidence, and pairwise feature similarity, where the peer confidence is the maximum class probability of the peer branch and the pairwise similarity is computed in the projected feature space. 
Therefore, each model learns from a reliability-aware peer feature target instead of a fixed average of the other branches.
The reliability-aware peer feature serves as the supervision target for feature-level knowledge distillation. To encourage semantic consistency across different event views, the projected feature of the current branch is aligned with its peer feature target using the cosine similarity loss:
\begin{equation}
\mathcal{L}_{F}^{m}=1-\operatorname{Cos}\left(\mathbf{q}_{m},\tilde{\mathbf{q}}_{m}\right),
\label{eq:feature_kd_loss}
\end{equation}
where $\mathbf{q}_{m}$ and $\tilde{\mathbf{q}}_{m}$ are the projected feature and the corresponding peer feature target, respectively, and $\operatorname{Cos}(\cdot,\cdot)$ denotes the cosine similarity. By maximizing the feature similarity, the proposed feature-level distillation encourages different branches to learn semantically consistent yet complementary representations across heterogeneous event views.

\subsection{Dynamic Hierarchical Model Merging}

Although collaborative multi-view learning enables the three branches to acquire complementary knowledge from different event representations, the resulting models remain independently optimized and therefore cannot fully exploit the shared knowledge accumulated during training. 
To obtain a unified and more robust model, we introduce a Dynamic Hierarchical Model Merging module, which progressively consolidates model parameters in a hierarchical manner.

Specifically, the proposed module consists of two stages. 
The first stage, referred to as dynamic historical model merging, dynamically integrates historical checkpoints within each branch through learnable weight averaging, producing a robust branch model that captures complementary optimization trajectories throughout training. 
The second stage, termed frequency-aware model merging, further integrates the aggregated models from different branches in the frequency domain, where shared transferable knowledge and branch-specific expertise are selectively preserved to construct the final unified model.

The dynamic historical model merging stage aggregates temporal checkpoints after collaborative multi-view training. 
It is optimized separately from the preceding Collaborative Multi-View Learning module. 
The historical merging coefficients are then learned while all checkpoint parameters remain fixed. Given $K$ historical checkpoints $\{\phi_i\}_{i=1}^{K}$ sampled along the training trajectory, the dynamically aggregated model is defined as:
\begin{equation}
\phi_{\mathrm{hist}}
=
\sum_{i=1}^{K}\beta_i\phi_i,
\quad
\sum_{i=1}^{K}\beta_i=1,\ \beta_i\ge 0.
\label{eq:historical_merge}
\end{equation}
The coefficients $\{\beta_i\}_{i=1}^{K}$ are trainable parameters.

The frequency-aware model merging stage then integrates the branch models in the parameter space. 
Specifically, for each branch $m\in\{f,s,v,hist\}$, let $\theta_{0}$ denote the initial parameters and $\theta_m$ denote the trained parameters of branch $m$. 
The task vector is defined as 
$\tau_m=\theta_m-\theta_{0}$. 
Each task vector is reshaped into a two-dimensional parameter map and transformed to the frequency domain. 
We denote the frequency-domain high-pass mask as $\mathbf{M}_{\mathrm{hp}}$, which suppresses low-frequency components that tend to contain branch-specific interference. 
Then the inverse transform produces a high-frequency shared component:
\begin{equation}
\begin{aligned}
\tau_{m,H}
&=
\mathcal{F}^{-1}
\left(
\mathbf{M}_{\mathrm{hp}}\odot\mathcal{F}(\tau_m)
\right),\\
\tau_{m,L}
&=
\tau_m-\tau_{m,H},
\end{aligned}
\label{eq:frequency_split}
\end{equation}
where $\mathcal{F}$ and $\mathcal{F}^{-1}$ denote the two-dimensional Fourier transform and inverse transform, and $\odot$ denotes element-wise multiplication. After the inverse transform, the filtered map is reshaped back to the original parameter-vector form. Specifically, $\tau_{m,H}$ has the same dimensionality as $\tau_m$ and represents the high-frequency transferable component of branch $m$, which is used for shared merging. $\tau_{m,L}$ is the complementary low-frequency residual, which retains branch-specific information for expert extraction.
The high-frequency components are merged as shared transferable knowledge:
\begin{equation}
\theta_{\mathrm{merge}}=\theta_0+\sum_{m\in\mathcal{M}}\lambda_m\tau_{m,H},
\label{eq:shared_merge}
\end{equation}
where $\lambda_m$ denotes the merging coefficient estimated from the normalized magnitude of the task vector. We extract a sparse branch expert from $\tau_{m,L}$ by retaining the top-magnitude entries.
\begin{equation}
\begin{aligned}
e_m
&=
\operatorname{TopK}(\tau_{m,L};d),\\
\theta_{\mathrm{out},m}
&=
\theta_{\mathrm{merge}}
+
\mu_m e_m,
\end{aligned}
\label{eq:branch_expert}
\end{equation}
where $d$ is the expert density and $\mu_m$ is the rescaling factor for the branch expert.

\subsection{Optimization Objective}
Since the proposed framework is optimized in two sequential stages, the corresponding objective functions are formulated separately for each stage. Specifically, the Collaborative Multi-View Learning module is first optimized to learn complementary representations across heterogeneous event views through supervised classification and mutual distillation. 
Based on the resulting branch models, the Dynamic Hierarchical Model Merging module is then optimized to progressively aggregate historical checkpoints and merge multi-view model parameters into a unified model. The optimization objectives for the two stages are described as follows.

\noindent
\textbf{Optimization of Collaborative Multi-View Learning.}
Besides two mutual distillation losses $\mathcal{L}_D^m$ and $\mathcal{L}_F^m$, the three RGB-Event branches are trained jointly with supervised classification loss $\mathcal{L}_C^m$ and relation-preserving feature alignment loss $\mathcal{L}_R^m$.
Specifically, for branch $m$, they are defined as
\begin{equation}
\begin{aligned}
\mathcal{L}_C^m
&=
-\sum_{c=1}^{C}y_c\log p_{m,c}, ~~~
\mathcal{L}_R^m&=
\left\|\mathbf{A}(\mathbf{q}_m)-\mathbf{A}(\tilde{\mathbf{q}}_m)\right\|_2^2,
\end{aligned}
\label{eq:loss_terms}
\end{equation}
where $\mathbf{A}(\cdot)$ denotes the normalized mini-batch feature similarity matrix. 
The overall objective of collaborative multi-view learning is
\begin{equation}
\mathcal{L}_{\mathrm{CML}}
=
\sum_{m\in\mathcal{M}}
\left(
\lambda_C\mathcal{L}_C^m
+
\lambda_D\mathcal{L}_D^m
+
\lambda_F\mathcal{L}_F^m
+
\lambda_R\mathcal{L}_R^m
\right).
\label{eq:overall_loss}
\end{equation}
In practice, the mutual distillation terms are enabled after warm-up epochs, reducing unstable peer supervision in early training.

\noindent
\textbf{Optimization of Dynamic Hierarchical Model Merging.}
In the second stage, the branch parameters obtained from collaborative multi-view learning are kept fixed. The dynamic historical model merging coefficients are optimized with the supervised classification loss:
\begin{equation}
\mathcal{L}_{\mathrm{DHM}}
=
\sum_{m\in\mathcal{M}}
\omega_m^{\mathrm{hist}}
\mathcal{L}_{C}^{m}(\phi_{\mathrm{hist}}),
\label{eq:historical_merge_objective}
\end{equation}
where $\omega_m^{\mathrm{hist}}$ is the branch weight used for coefficient learning. The optimization searches for the coefficient values that minimize the above loss. The frequency-aware model merging stage is then performed in parameter space, without introducing an additional end-to-end training loss.

Overall, the proposed framework consists of three sequential components, where the Multi-View Event Encoding module provides a unified representation interface for heterogeneous event data, while the primary contributions lie in the subsequent Collaborative Multi-View Learning and Dynamic Hierarchical Model Merging modules. 
The former promotes complementary representation learning through collaborative mutual distillation across multiple event views, whereas the latter hierarchically aggregates both historical optimization knowledge and cross-view model parameters to obtain a unified and more generalizable model. 
By jointly exploring complementarity at both the representation level and the parameter level, the proposed framework effectively bridges heterogeneous event representations and achieves robust RGB-Event representation learning.

\section{Experiments} \label{sec::experiments}

We evaluate the proposed method on two public RGB-Event recognition benchmarks, PokerEvent~\citep{wang2025sstformer} and HARDVS~\citep{wang2024hardvs}.

\subsection{Comparison on Public Benchmark Datasets}

\noindent $\bullet$ \textbf{Results on PokerEvent Dataset~\citep{wang2025sstformer}.} As illustrated in Table~\ref{tab:Poker}, we compare the proposed method with recent state-of-the-art (SOTA) approaches on the PokerEvent dataset. Our model achieves the highest Top-1 accuracy of 58.73\%, outperforming the previous best method ViP-HMNN~\citep{ji2025vip}. Compared with other competitive recent methods, such as TSCFormer~\citep{wang2025unleashing}, SAFE~\citep{li2025semantic}, and MMHCO-HAR~\citep{wang2025human}, our approach also shows consistent performance improvements. Notably, these methods cover transformer-based, VLM-based, and multimodal HCO-based frameworks, indicating that our model remains competitive against diverse advanced recognition paradigms. These results demonstrate the effectiveness of the proposed modules and validate their generalization ability for human activity recognition in challenging poker-event scenarios.

\begin{table}[!t]
\centering
\caption{Comparison of recognition performance and computational efficiency on the PokerEvent~\citep{wang2025sstformer} dataset. Efficiency metrics include the number of parameters, FLOPs, training throughput, and inference latency. ``-'' indicates that the corresponding result is not reported. The best three Top-1 results are highlighted in \textcolor{red}{\textbf{red}}, \textcolor{blue}{\textbf{blue}}, and \textbf{bold}, respectively.}
\label{tab:Poker}
\label{tab:computation_cost}
\small\setlength{\tabcolsep}{2pt}
\begin{tabular}{@{}clllrrrrr@{}} 
\toprule
\textbf{\#No.} &
\textbf{Methods} &
\textbf{Source} &
\textbf{Backbone} &
\makecell{\textbf{Params}\\\textbf{(M)}} &
\makecell{\textbf{FLOPs}\\\textbf{(G)}} &
\makecell{\textbf{Train}\\\textbf{(clip/s)}} &
\makecell{\textbf{Infer.}\\\textbf{(ms)}} &
\textbf{Top-1} \\
\midrule
01 & C3D
& ICCV 2015 & 3D CNN
& 34.8 & 21.7 & 12.65 & 35.69 & 51.76 \\
02 & TSM
& ICCV 2019 & ResNet50
& 24.3 & 65.0 & 10.31 & 30.58 & 55.43 \\
03 & X3D
& CVPR 2020 & ResNet
& 11.1 & 48.4 & 27.29 & 16.34 & 51.75 \\
04 & TimeSformer
& ICML 2021 & ViT
& 121.4 & 53.6 & 2.83 & 114.78 & 55.69 \\
05 & ACTION-Net
& CVPR 2021 & ResNet50
& 27.9 & 17.3 & 3.30 & 75.34 & 54.29 \\
06 & MVIT
& CVPR 2022 & ViT
& 52.0 & 10.2 & 3.35 & 319.65 & 55.02 \\
07 & V-SwinTrans
& CVPR 2022 & Swin-Former
& 88.1 & 282.0 & 11.91 & 214.64 & 54.17 \\
08 & Vision Mamba
& ICML 2024 & Mamba
& 25.6 & 18.1 & 9.25 & 35.29 & 56.50 \\
09 & SAFE
& PR 2024 & VLM
& 243.2 & 943.9 & 2.03 & 181.09 & 57.64 \\
10 & TSCFormer
& MIR 2025 & TSCFormer
& 54.5 & 34.5 & 54.39 & 68.83 & \textbf{57.70} \\
11 & SSTFormer
& TCDS 2025 & SNN-Former
& 336.5 & 1104.8 & 2.38 & 138.74 & 54.74 \\
12 & EFV++
& TMM 2025 & Former-GNN
& 160.8 & 45.6 & 10.71 & 24.28 & 55.40 \\
13 & ViP-HMNN
& Info. Fusion 2026 & ResNet50
& - & - & - & 90.0 & \textcolor{blue}{\textbf{58.27}} \\
14 & MMHCO-HAR
& IJCV 2026 & Multi-modal HCO
& 76.1 & 61.7 & - & - & 57.40 \\
\midrule
15 & Ours
& - & ResNet50
& 186.2 & 224.2 & 9.40 & 78.37
& \textcolor{red}{\textbf{58.73}} \\
\bottomrule
\end{tabular}
\end{table}

\noindent $\bullet$ \textbf{Results on HARDVS Dataset~\citep{wang2024hardvs}.} As shown in Table~\ref{tab:HARDVS}, our method achieves the best overall performance on the HARDVS dataset, with a Top-1 accuracy of 53.28\% and a Top-5 accuracy of 63.57\%. The strong performance on both metrics demonstrates that the proposed method can effectively integrate complementary information from heterogeneous event representations and generate reliable predictions across multiple candidate categories. Since HARDVS contains diverse human activities captured under varying viewpoints, illumination conditions, and motion speeds, these results further validate the robustness and generalization ability of the proposed framework in challenging real-world scenarios.

\begin{table}[!t]
\centering
\caption{Results on the HARDVS~\citep{wang2024hardvs} dataset. The best three results are highlighted in \textcolor{red}{\textbf{red}}, \textcolor{blue}{\textbf{blue}} and \textbf{bold}.}
\label{tab:HARDVS}
\resizebox{0.85\linewidth}{!}{
\begin{tabular}{c|l|l|l|c|c}    
\toprule
\textbf{\#No.} & \textbf{Methods} & \textbf{Source} & \textbf{Backbone} & \textbf{Top-1} & \textbf{Top-5}  \\  
\midrule
01& C3D                 & ICCV 2015 & 3D CNN & 50.88 & 56.51 \\
02& SlowFast   & ICCV 2019 & ResNet50 & 46.54 & 54.76 \\
03& TSM                       & ICCV 2019 & ResNet50 & 52.58 & 62.12 \\
04& X3D             & CVPR 2020 & ResNet & 47.38 & 51.42 \\
05& TimeSformer       & ICML 2021 & ViT & 51.57 & 58.48 \\
06& ACTION-Net            & CVPR 2021 & ResNet50 & 46.85 & 56.19 \\
07& V-SwinTrans             & CVPR 2022 & Swin-Former & 51.91 & 59.11 \\
08& ESTF                  & AAAI 2024 & ResNet18 & 49.93 & 55.77 \\
09& SCNN-MST           & TCDS 2025 & SCNN-Former & 49.40 & 55.55 \\
10& TSCFormer         & MIR 2025 & TSCFormer & \textbf{53.04} & \textcolor{blue}{\textbf{62.67}} \\
11& HoloEv-Net             & arXiv 2026 & MobileNetV3 & 52.69 & \textbf{62.56} \\
12& EIS-HAR            & CVPR 2026 & Swin-Former & \textcolor{blue}{\textbf{53.21}} & 60.45 \\
13& ViP-HMNN                   & Info. Fusion 2026 & ResNet50 & 51.88 & 58.98 \\
\midrule
14& Ours & - & ResNet50 & \textcolor{red}{\textbf{53.28}} & \textcolor{red}{\textbf{63.57}} \\
\bottomrule 
\end{tabular}
}
\end{table}

\subsection{Ablation Study}
To systematically assess the contribution of each component in the proposed method, we performed a comprehensive ablation study on the PokerEvent dataset.

\noindent $\bullet$ \textbf{Component Analysis.~} 
As shown in Table~\ref{tab:ablation_component}, we evaluate the contributions of the key components, including Collaborative Multi-View Learning (CML) and Dynamic Hierarchical Model Merging (DHMM), to the model accuracy on the PokerEvent dataset. The Baseline denotes the setting where the three event-view branches are trained independently without mutual distillation or model merging, and we report the best-performing branch. Compared with this Baseline, introducing CML improves the Top-1 accuracy from 56.48\% to 58.28\%, yielding a gain of 1.80\%. This improvement indicates that online peer supervision across event frame, event stream, and event voxel views can transfer complementary information and produce more discriminative branch representations than isolated training. After further applying DHMM, the full model reaches 58.73\%, bringing an additional gain of 0.45\% over CML alone and a total gain of 2.25\% over the Baseline. These results suggest that CML and DHMM play complementary roles: CML improves branch learning during training, while DHMM consolidates the transferable parameters after training.

\noindent $\bullet$ \textbf{Event-View Complementarity Analysis.~} 
As illustrated in Table~\ref{tab:ablation_event_view}, we evaluate the contribution of each event view on the PokerEvent dataset. When each view is trained and evaluated independently, the event voxel view achieves the best Top-1 accuracy of 56.48\%, outperforming the event frame view with 55.81\% and the event stream view with 55.54\%. This suggests that the voxel representation provides a stronger standalone spatio-temporal encoding, while the frame and stream views still preserve complementary spatial and asynchronous cues. After combining all three event views with CML, the accuracy increases to 58.28\%, bringing a 1.80\% gain over the strongest single-view result. These results demonstrate that the three event views are complementary and that mutual distillation effectively integrates their knowledge into more discriminative representations.

\begin{table}[!t]
\centering
\caption{Ablation study on the PokerEvent dataset. CML and DHMM denote Collaborative Multi-View Learning and Dynamic Hierarchical Model Merging, respectively. DHIS and FA are the dynamic historical model merging and frequency-aware model merging components of DHMM. Symbols ``$\times$'' and ``$\checkmark$'' indicate exclusion and inclusion, respectively.}
\label{tab:ablation_combined}
\label{tab:ablation_component}
\label{tab:ablation_event_view}
\label{tab:ablation_merge}
\resizebox{0.74\linewidth}{!}{
\begin{tabular}{l|ccc|c|cc|c}
\toprule
\multirow{2}{*}{Configuration}
& \multicolumn{3}{c|}{Event View}
& \multirow{2}{*}{CML}
& \multicolumn{2}{c|}{DHMM}
& \multirow{2}{*}{Top-1} \\
\cmidrule(lr){2-4}\cmidrule(lr){6-7}
& Frame & Stream & Voxel & & DHIS & FA & \\
\midrule
Frame only
& $\checkmark$ & $\times$ & $\times$
& $\times$ & $\times$ & $\times$ & 55.81 \\
Stream only
& $\times$ & $\checkmark$ & $\times$
& $\times$ & $\times$ & $\times$ & 55.54 \\
Voxel only 
& $\times$ & $\times$ & $\checkmark$
& $\times$ & $\times$ & $\times$ & 56.48 \\
\midrule
CML only
& $\checkmark$ & $\checkmark$ & $\checkmark$
& $\checkmark$ & $\times$ & $\times$ & 58.28 \\
\midrule
CML + DHIS
& $\checkmark$ & $\checkmark$ & $\checkmark$
& $\checkmark$ & $\checkmark$ & $\times$ & 58.43 \\
CML + FA
& $\checkmark$ & $\checkmark$ & $\checkmark$
& $\checkmark$ & $\times$ & $\checkmark$ & 58.62 \\
Ours
& $\checkmark$ & $\checkmark$ & $\checkmark$
& $\checkmark$ & $\checkmark$ & $\checkmark$ & \textbf{58.73} \\
\bottomrule
\end{tabular}
}
\end{table}

\begin{figure}[!t]
\centering
\includegraphics[width=\textwidth]{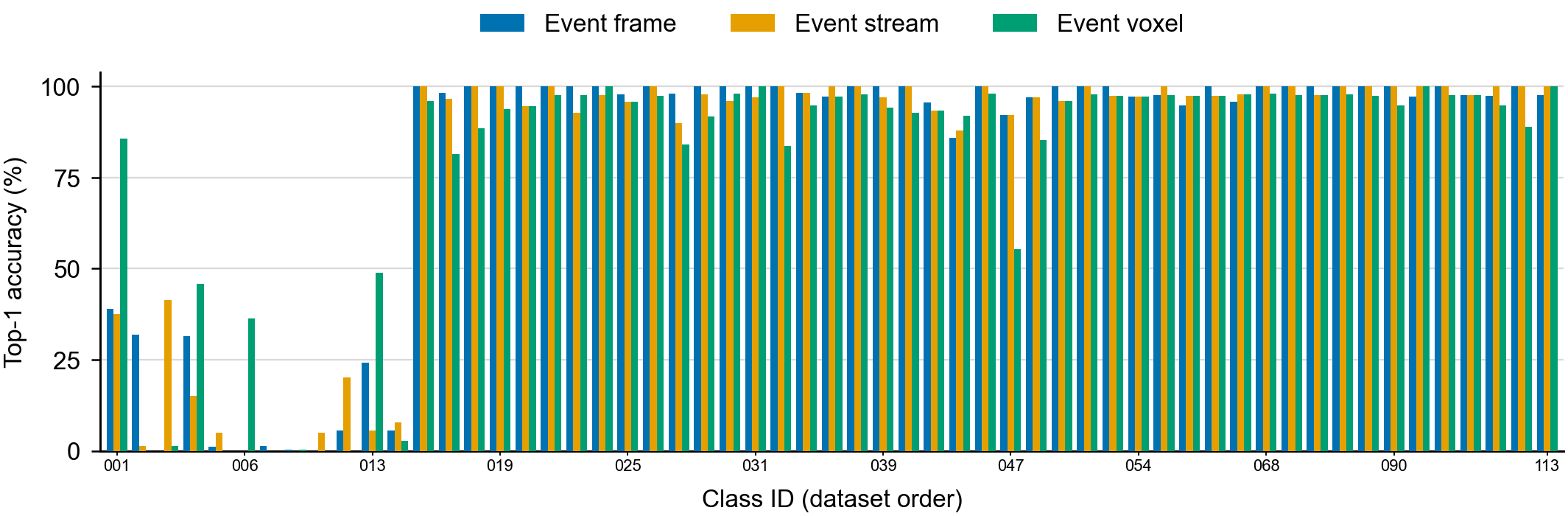}
\caption{Class-wise Top-1 accuracy of the event-frame, event-stream, and event-voxel branches on PokerEvent in the original dataset order. For clarity, we omit 55 classes scored at 100\% by all three branches and two scored at 0\%, leaving 57 classes.}
\label{fig:poker_classwise_view}
\end{figure}

\noindent $\bullet$ \textbf{Class-wise Modality Complementarity Analysis.} Fig.~\ref{fig:poker_classwise_view} visualizes how the three event-modality branches behave across PokerEvent categories. Among the 57 displayed classes, the branches produce different accuracies on 55 classes. Moreover, 28 classes have a unique best-performing branch: the event-frame, event-stream, and event-voxel branches lead on 15, 8, and 5 classes, respectively. This distribution reveals class-dependent complementarity rather than consistent dominance by a single event representation.

\subsection{Efficiency Analysis} \label{sec::EfficiencyAnalysis} 
To evaluate the efficiency benefit of the model merging design, we compare the original inference setting without model merging and the merged model under the same testing configuration. 

\begin{wrapfigure}[8]{r}{0.45\textwidth}
\centering
\small
\begin{tabular}{lcc}
\toprule
Setting
& \makecell{Total GPU\\Memory (MB)}
& \makecell{Peak Allocation\\(MB)} \\
\midrule
Before 
& 5642 & 3970 \\
After  
& 3442 & 1768 \\
\bottomrule
\end{tabular}
\caption{Memory usage analysis.} 
\label{tab:efficiency_memory}
\end{wrapfigure}

As shown in Table~\ref{tab:efficiency_memory}, model merging reduces the total GPU memory footprint from 5642 MB to 3442 MB, corresponding to a 39.0\% reduction. More importantly, the peak model allocation decreases from 3970 MB to 1768 MB, achieving a 55.5\% reduction. 
These results indicate that the proposed merging strategy preserves the multi-view knowledge learned from the three event representations while substantially lowering the memory requirement during deployment.


\section{Conclusion} \label{sec::conclusion}
We presented a multi-view approach to RGB-Event representation learning that considers complementarity in both representation space and parameter space. Mutual distillation promotes knowledge exchange among event-frame, event-stream, and event-voxel representations, while dynamic hierarchical merging integrates historical optimization knowledge and cross-view model parameters. Results on PokerEvent and HARDVS demonstrate the effectiveness of this formulation for fine-grained recognition and human activity recognition. These findings support the joint consideration of collaborative representation learning and parameter consolidation.

\section*{AI Use Statement}
Generative AI tools (large language models) were employed in this work to assist with manuscript composition and language polishing. We did not use generative AI tools for other research activities, such as research conception, experimental implementation, coding, figure generation, or data analysis. We have carefully reviewed all AI-assisted text and bear full responsibility for the final content of this paper.

\label{page:main_end}
\bibliographystyle{other/iclr2027_conference}
\bibliography{other/reference}

@article{wang2025sstformer,
  title={Sstformer: Bridging spiking neural network and memory support transformer for frame-event based recognition},
  author={Wang, Xiao and Rong, Yao and Wu, Zongzhen and Zhu, Lin and Jiang, Bo and Tang, Jin and Tian, Yonghong},
  journal={IEEE Transactions on Cognitive and Developmental Systems},
  year={2025},
  publisher={IEEE}
}

@article{wang2025unleashing,
  title={Unleashing the power of cnn and transformer for balanced rgb-event video recognition},
  author={Wang, Xiao and Rong, Yao and Wang, Shiao and Chen, Yuan and Wu, Zhe and Jiang, Bo and Tian, Yonghong and Tang, Jin},
  journal={Machine Intelligence Research},
  volume={22},
  number={6},
  pages={1031--1047},
  year={2025},
  publisher={Springer}
}

@article{wang2025human,
  title={Human Activity Recognition using RGB-Event based Sensors: A Multi-modal Heat Conduction Model and A Benchmark Dataset},
  author={Wang, Shiao and Wang, Xiao and Jiang, Bo and Zhu, Lin and Li, Guoqi and Wang, Yaowei and Tian, Yonghong and Tang, Jin},
  journal={arXiv preprint arXiv:2504.05830},
  year={2025}
}

@article{ji2025vip,
  title={ViP-HMNN: A Visual Pathway-Inspired Hybrid Neural Network Incorporated with In-Memory Computing for Object Recognition},
  author={Ji, Xiaoyue and Han, Yifeng and Lai, Chun Sing and Yu, Fangwen and Dong, Zhekang},
  journal={Information Fusion},
  pages={104086},
  year={2025},
  publisher={Elsevier}
}

@article{ilharco2022editing,
  title={Editing models with task arithmetic},
  author={Ilharco, Gabriel and Ribeiro, Marco Tulio and Wortsman, Mitchell and Gururangan, Suchin and Schmidt, Ludwig and Hajishirzi, Hannaneh and Farhadi, Ali},
  journal={arXiv preprint arXiv:2212.04089},
  year={2022}
}

@article{du2024parameter,
  title={Parameter competition balancing for model merging},
  author={Du, Guodong and Lee, Junlin and Li, Jing and Jiang, Runhua and Guo, Yifei and Yu, Shuyang and Liu, Hanting and Goh, Sim K and Tang, Ho-Kin and He, Daojing and others},
  journal={Advances in Neural Information Processing Systems},
  volume={37},
  pages={84746--84776},
  year={2024}
}

@article{lu2024twin,
  title={Twin-merging: Dynamic integration of modular expertise in model merging},
  author={Lu, Zhenyi and Fan, Chenghao and Wei, Wei and Qu, Xiaoye and Chen, Dangyang and Cheng, Yu},
  journal={Advances in Neural Information Processing Systems},
  volume={37},
  pages={78905--78935},
  year={2024}
}

@article{marczak2025no,
  title={No task left behind: Isotropic model merging with common and task-specific subspaces},
  author={Marczak, Daniel and Magistri, Simone and Cygert, Sebastian and Twardowski, Bart{\l}omiej and Bagdanov, Andrew D and Van De Weijer, Joost},
  journal={arXiv preprint arXiv:2502.04959},
  year={2025}
}

@inproceedings{zheng2025free,
  title={Free-merging: Fourier transform for efficient model merging},
  author={Zheng, Shenghe and Wang, Hongzhi},
  booktitle={Proceedings of the IEEE/CVF International Conference on Computer Vision},
  pages={3863--3873},
  year={2025}
}

@article{lee2025adarank,
  title={Adarank: Adaptive rank pruning for enhanced model merging},
  author={Lee, Chanhyuk and Choi, Jiho and Lee, Chanryeol and Kim, Donggyun and Hong, Seunghoon},
  journal={arXiv preprint arXiv:2503.22178},
  year={2025}
}

@inproceedings{liu2026globa,
  title={GLOBA: Rethinking Parameter Conflicts in Model Merging},
  author={Liu, Zehao and Li, Kun and Zhou, Wei},
  booktitle={Proceedings of the AAAI Conference on Artificial Intelligence},
  volume={40},
  number={28},
  pages={23954--23962},
  year={2026}
}

@inproceedings{zhang2018deep,
  title={Deep mutual learning},
  author={Zhang, Ying and Xiang, Tao and Hospedales, Timothy M and Lu, Huchuan},
  booktitle={Proceedings of the IEEE conference on computer vision and pattern recognition},
  pages={4320--4328},
  year={2018}
}

@inproceedings{guo2020online,
  title={Online knowledge distillation via collaborative learning},
  author={Guo, Qiushan and Wang, Xinjiang and Wu, Yichao and Yu, Zhipeng and Liang, Ding and Hu, Xiaolin and Luo, Ping},
  booktitle={Proceedings of the IEEE/CVF conference on computer vision and pattern recognition},
  pages={11020--11029},
  year={2020}
}

@inproceedings{shi2025competitive,
  title={Competitive Distillation: A Simple Learning Strategy for Improving Visual Classification},
  author={Shi, Daqian and Diao, Xiaolei and Chen, Xu and John, C{\'e}dric M},
  booktitle={Proceedings of the IEEE/CVF International Conference on Computer Vision},
  pages={2981--2990},
  year={2025}
}

@inproceedings{wang2024hardvs,
  title={Hardvs: Revisiting human activity recognition with dynamic vision sensors},
  author={Wang, Xiao and Wu, Zongzhen and Jiang, Bo and Bao, Zhimin and Zhu, Lin and Li, Guoqi and Wang, Yaowei and Tian, Yonghong},
  booktitle={Proceedings of the AAAI conference on artificial intelligence},
  volume={38},
  number={6},
  pages={5615--5623},
  year={2024}
}

@article{chen2025retain,
  title={Retain, blend, and exchange: A quality-aware spatial-stereo fusion approach for event stream recognition},
  author={Chen, Lan and Li, Dong and Wang, Xiao and Shao, Pengpeng and Zhang, Wei and Wang, Yaowei and Tian, Yonghong and Tang, Jin},
  journal={IEEE Transactions on Multimedia},
  volume={27},
  pages={8926--8939},
  year={2025},
  publisher={IEEE}
}

@article{li2025semantic,
  title={Semantic-aware frame-event fusion based pattern recognition via large vision--language models},
  author={Li, Dong and Jin, Jiandong and Zhang, Yuhao and Zhong, Yanlin and Wu, Yaoyang and Chen, Lan and Wang, Xiao and Luo, Bin},
  journal={Pattern Recognition},
  volume={158},
  pages={111080},
  year={2025},
  publisher={Elsevier}
}

@inproceedings{wang2021evdistill,
  title={Evdistill: Asynchronous events to end-task learning via bidirectional reconstruction-guided cross-modal knowledge distillation},
  author={Wang, Lin and Chae, Yujeong and Yoon, Sung-Hoon and Kim, Tae-Kyun and Yoon, Kuk-Jin},
  booktitle={Proceedings of the IEEE/CVF Conference on Computer Vision and Pattern Recognition},
  pages={608--619},
  year={2021}
}

@inproceedings{li2024object,
  title={Object-centric cross-modal feature distillation for event-based object detection},
  author={Li, Lei and Linger, Alexander and Millh{\"a}usler, Mario and Tsiminaki, Vagia and Li, Yuanyou and Dai, Dengxin},
  booktitle={2024 IEEE International Conference on Robotics and Automation (ICRA)},
  pages={15440--15447},
  year={2024},
  organization={IEEE}
}

@inproceedings{huo2024c2kd,
  title={C2kd: Bridging the modality gap for cross-modal knowledge distillation},
  author={Huo, Fushuo and Xu, Wenchao and Guo, Jingcai and Wang, Haozhao and Guo, Song},
  booktitle={Proceedings of the IEEE/CVF Conference on Computer Vision and Pattern Recognition},
  pages={16006--16015},
  year={2024}
}

@article{liu2024frequency,
  title={Frequency-Decoupled Cross-Modal Knowledge Distillation},
  author={Liu, Junhong and Zhang, Yuan and Huang, Tao and Wang, Haozhao and Xu, Wenchao and Zhang, Shanghang and Yang, Renyu},
  year={2024}
}

@article{yadav2023ties,
  title={Ties-merging: Resolving interference when merging models},
  author={Yadav, Prateek and Tam, Derek and Choshen, Leshem and Raffel, Colin A and Bansal, Mohit},
  journal={Advances in neural information processing systems},
  volume={36},
  pages={7093--7115},
  year={2023}
}

@inproceedings{davari2024model,
  title={Model breadcrumbs: Scaling multi-task model merging with sparse masks},
  author={Davari, MohammadReza and Belilovsky, Eugene},
  booktitle={European Conference on Computer Vision},
  pages={270--287},
  year={2024},
  organization={Springer}
}

@article{huang2024emr,
  title={Emr-merging: Tuning-free high-performance model merging},
  author={Huang, Chenyu and Ye, Peng and Chen, Tao and He, Tong and Yue, Xiangyu and Ouyang, Wanli},
  journal={Advances in Neural Information Processing Systems},
  volume={37},
  pages={122741--122769},
  year={2024}
}

@inproceedings{yang2024adamerging,
  title={Adamerging: Adaptive model merging for multi-task learning},
  author={Yang, Enneng and Wang, Zhenyi and Shen, Li and Liu, Shiwei and Guo, Guibing and Wang, Xingwei and Tao, Dacheng},
  booktitle={International Conference on Learning Representations},
  volume={2024},
  pages={22743--22763},
  year={2024}
}
\appendix
\section{Experimental Setup}
\subsection{Datasets and Evaluation Metric}
\noindent $\bullet$ \textbf{PokerEvent Dataset}~\citep{wang2025sstformer} is a dual-modality dataset focusing on the recognition of character patterns in poker cards. It contains 114 classes and includes 27,102 aligned RGB-Event pairs recorded using a DAVIS 346 event camera, which simultaneously outputs RGB frames and event streams. This benchmark requires fine-grained recognition of visually similar poker patterns under dynamic event responses. The dataset is divided into training, validation, and testing subsets, with 16,216, 2,687, and 8,199 samples, respectively.

\noindent $\bullet$ \textbf{HARDVS Dataset}~\citep{wang2024hardvs} is a large-scale human activity recognition dataset collected using a DAVIS 346 event camera. It encompasses 300 categories of daily human activities, such as drinking, riding a bike, and sitting down, and includes 107,646 RGB-Event video sequences. The dataset covers challenging real-world factors, including multiple viewing angles, illumination conditions, and motion speeds, making it suitable for evaluating robustness in complex activity scenarios. It is divided into training, validation, and testing subsets, with 64,526, 10,734, and 32,386 samples, respectively. 

For evaluation, we follow the commonly used accuracy metrics of each benchmark. On PokerEvent~\citep{wang2025sstformer}, we report \textbf{Top-1 accuracy}, which measures whether the highest-confidence prediction matches the ground-truth label. On HARDVS~\citep{wang2024hardvs}, we report both \textbf{Top-1 accuracy} and \textbf{Top-5 accuracy}, where Top-5 accuracy counts a prediction as correct if the ground-truth label appears among the five highest-confidence classes.

\subsection{Implementation Details}
We implement the proposed framework in PyTorch based on MMAction2 and conduct all experiments on NVIDIA RTX 3090 GPUs. Each RGB-Event branch adopts an ImageNet-pretrained ResNet50-based backbone. During training, the same multi-scale crop and horizontal flip are synchronously applied to the RGB input, while center cropping is used during evaluation. We use SGD with momentum $0.9$, weight decay $5\times10^{-5}$, and gradient clipping with a maximum norm of $20$. The distillation temperature is set to $T_d=4$. The classification and logit-distillation weights are $\lambda_C=0.6$ and $\lambda_D=0.1$, respectively, and logit-level mutual distillation is activated from epoch $6$. The model is trained for $30$ epochs with a batch size of $5$ and an initial learning rate of $8\times10^{-4}$. Feature-level distillation and relation-preserving distillation are activated from epoch $8$, with $\lambda_F=0.15$ and $\lambda_R=0.15$. For model merging, we merge the RGB-Event fusion and classification-head parameter groups. The frequency cutoff and sparse-expert density are set to $0.005$ and $0.01$, respectively. We use the checkpoints from epochs $\{5,10,15,20,25,30\}$ and optimize their coefficients on the training split.

\section{Ablation Study}
\noindent $\bullet$ \textbf{Model-Merging Strategy Analysis.~} 
As shown in Table~\ref{tab:ablation_merge}, we evaluate the individual and joint contributions of Dynamic Historical model merging (DHIS) and frequency-aware model merging (FA) after CML training. Using CML alone achieves 58.28\% Top-1 accuracy. When only DHIS is applied, the result increases to 58.43\%, showing that dynamic historical model merging provides a stable complementary solution. When only the FA is used, the result improves to 58.62\%, indicating that parameter merging can consolidate transferable knowledge from the multi-view networks. Combining DHIS with the FA achieves the best Top-1 accuracy of 58.73\%, demonstrating that the strategy provides complementary benefits.

\noindent $\bullet$ \textbf{Comparison with Model Merging Methods.~}
Table~\ref{tab:merging_method_comparison} compares our method with representative model merging approaches. Weight Averaging yields only 21.17\% Top-1 accuracy, suggesting that uniformly averaging heterogeneous branch parameters causes severe interference. Task Arithmetic\citep{ilharco2022editing} and AdaMerging~\citep{yang2024adamerging} achieve Top-1 accuracies from 55.23\% to 55.42\%. In contrast, our method reaches 58.73\%, exceeding the strongest competing result by 3.31\%. This consistent advantage supports the effectiveness of hierarchically combining learned historical aggregation with frequency aware cross view merging for heterogeneous event representations.

\begin{table}[!htbp]
\centering
\caption{Comparison with different model merging methods on the PokerEvent dataset.}
\label{tab:merging_method_comparison}

\resizebox{\linewidth}{!}{
\begin{tabular}{l|c|l|c}
\toprule
\textbf{Methods} & \textbf{Top-1} & \textbf{Methods} & \textbf{Top-1} \\
\midrule
Weight Averaging & 21.17 & Ties-Merging~\citep{yadav2023ties} & 55.32 \\
Task Arithmetic\citep{ilharco2022editing} & 55.23 & Breadcrumbs~\citep{davari2024model} & 55.40 \\
EMR-Merging~\citep{huang2024emr} & 54.73 & AdaMerging~\citep{yang2024adamerging} & 55.42 \\
Twin-Merging~\citep{lu2024twin} & 55.42 & Ours & \textbf{58.73} \\
\bottomrule
\end{tabular}
}
\end{table}

\noindent $\bullet$ \textbf{Effect of the Number of Historical Epochs.~}
Table~\ref{tab:dhis_epoch_number} studies the effect of the number of historical epochs used in DHIS. The model achieves its highest Top-1 accuracy of 58.73\% when 5 epochs are used, whereas the other settings remain within the range from 58.43\% to 58.53\%. In particular, increasing the number of epochs beyond 5 does not lead to further improvement, indicating that simply incorporating more historical checkpoints is not necessarily beneficial because later checkpoints may contain redundant or interfering optimization information. These results show that DHIS is generally stable across different numbers of historical epochs, while selecting 5 epochs provides the best balance between trajectory diversity and parameter compatibility.

\begin{table}[!htbp]
\centering
\caption{Effect of the number of epochs used in DHIS on the PokerEvent dataset.}
\label{tab:dhis_epoch_number}

\begin{tabular}{c|c|c|c}
\toprule
\textbf{\#Epochs} & \textbf{Top-1} & \textbf{\#Epochs} & \textbf{Top-1} \\
\midrule
2 & 58.53 & 6  & 58.43 \\
4 & 58.51 & 8  & 58.43 \\
5 & \textbf{58.73} & 10 & 58.46 \\
\bottomrule
\end{tabular}
\end{table}

\noindent $\bullet$ \textbf{Analysis of Hyperparameters in Model Merge.~} Table~\ref{tab:merge_hyperparameters} examines the sensitivity of model merging to the frequency cutoff and sparse expert density. The results remain close across all settings, with Top-1 accuracy staying around 58.6\%. The smallest tested frequency cutoff yields the strongest result, suggesting that retaining a broader set of frequency components benefits shared knowledge integration. For sparse expert extraction, a moderate density performs best, while overly sparse or dense experts provide no further benefit. These observations support the selected configuration as an effective balance between shared parameter integration and the preservation of branch specific information.

\begin{table}[!htbp]
\centering
\caption{Analysis of the frequency cutoff and sparse expert density in model merging on the PokerEvent dataset.}
\label{tab:merge_hyperparameters}

\begin{tabular}{c|c|c|c}
\toprule
\textbf{Frequency Cutoff} & \textbf{Top-1} & \textbf{Expert Density} & \textbf{Top-1} \\
\midrule
0.005  & \textbf{58.73} & 0.001     & 58.68 \\
0.050  & 58.68          & 0.010     & \textbf{58.73} \\
0.200  & 58.63          & 0.100     & 58.63 \\
0.500  & 58.64          & 0.500     & 58.65 \\
\bottomrule
\end{tabular}
\end{table}

\noindent $\bullet$ \textbf{Analysis of Distilling Strategies in Collaborative Multi-View Learning.~} Table~\ref{tab:ablation_distillation} evaluates how the three distillation objectives contribute to Collaborative Multi View Learning. Logit distillation provides stronger supervision than feature distillation when each objective is used alone. Adding relation preservation to feature distillation improves the result, which shows that maintaining pairwise feature structure complements direct feature alignment. The complete strategy combines prediction alignment, feature alignment, and relation preservation and achieves the best Top-1 accuracy of 58.28\%. This result demonstrates that the three objectives provide complementary supervision at different representation levels and work most effectively when optimized together.

\begin{table}[!htbp]
\centering
\caption{Ablation of distillation strategies on the PokerEvent dataset.}
\label{tab:ablation_distillation}

\begin{tabular}{c|ccc|c}
\toprule
\textbf{Setting} & \textbf{Logit KD} & \textbf{Feature KD} & \textbf{Relation} & \textbf{Top-1} \\
\midrule
(a) & $\checkmark$ & $\times$ & $\times$ & 56.99 \\
(b) & $\times$ & $\checkmark$ & $\times$ & 56.41 \\
(c) & $\times$ & $\checkmark$ & $\checkmark$ & 56.92 \\
(d) & $\checkmark$ & $\checkmark$ & $\checkmark$ & \textbf{58.28} \\
\bottomrule
\end{tabular}
\end{table}

\section{Efficiency Analysis} 
As shown in Table~\ref{tab:computation_cost}, we compare representative methods in terms of model size, computational cost, training throughput, inference latency, and precision on the PokerEvent dataset. Our method achieves the highest precision of 58.73\%, outperforming all compared baselines, including recent RGB-Event and video recognition models such as TSCFormer~\citep{wang2025unleashing} and EFV++~\citep{chen2025retain}. In terms of efficiency, our model requires 186.2M parameters and 224.2G FLOPs, which is higher than lightweight single-backbone methods. This overhead mainly comes from the multi-view design, where event frame, event stream, and event voxel representations are jointly exploited to capture complementary spatio-temporal cues. Nevertheless, the proposed model still maintains a practical training throughput of 9.40 clips/s and an inference latency of 78.37 ms. Compared with high-capacity alternatives such as SAFE~\citep{li2025semantic} and SSTformer~\citep{wang2025sstformer}, our method uses fewer FLOPs, runs faster during both training and inference, and achieves substantially better precision. These results indicate that the proposed mutually-distilled multi-view learning and model merging strategy offers a favorable accuracy-efficiency trade-off: it improves recognition performance by integrating complementary event views, while keeping the additional computational and deployment cost within a reasonable range.

\begin{figure}[!t]
\centering
\includegraphics[width=\linewidth]{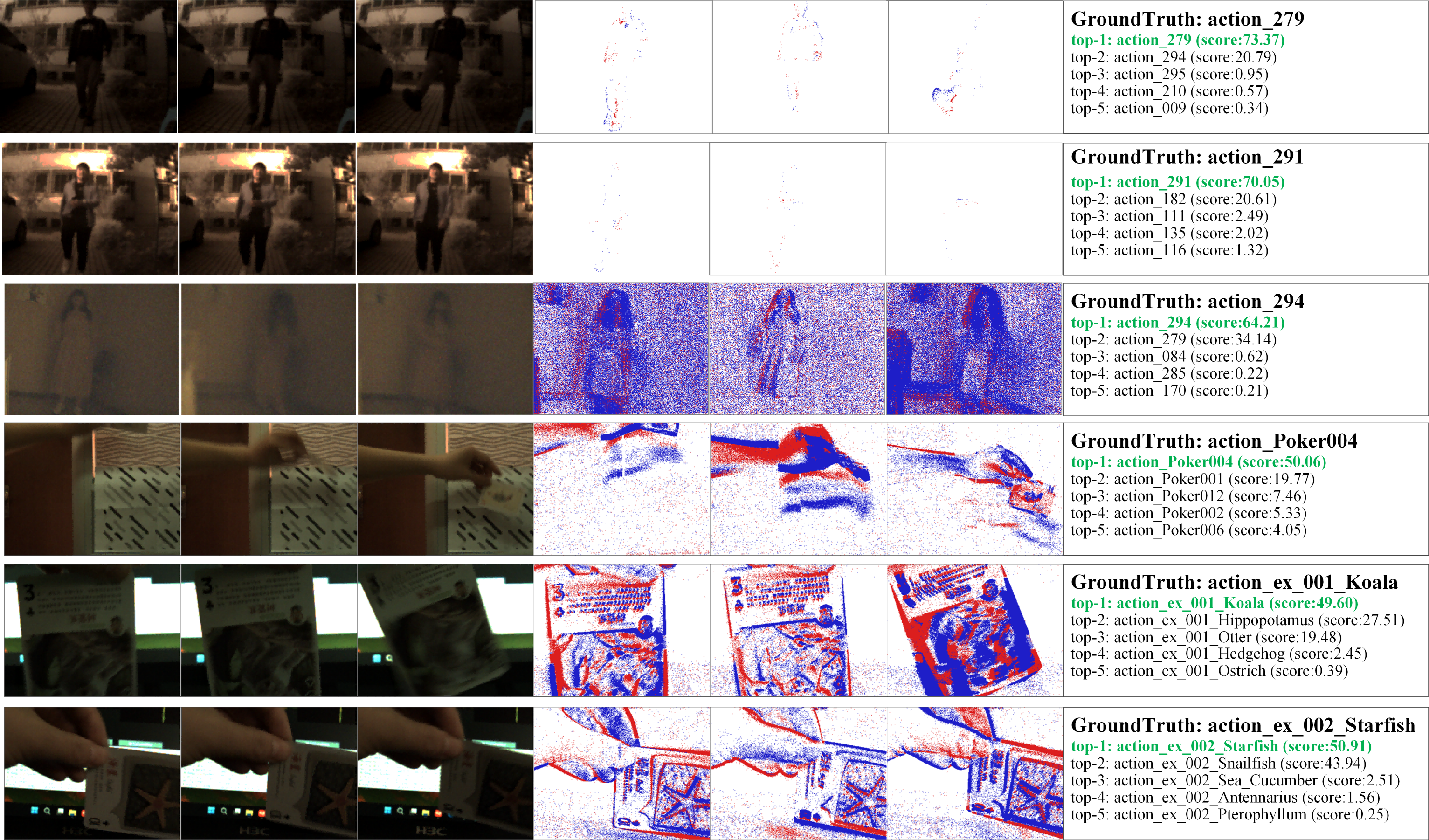}
\caption{Visualization of Top-5 accuracy on the HARDVS and the PokerEvent dataset.} 
\label{fig:Vis_Top-5}
\end{figure}

\section{Limitation Analysis} 
Although the proposed framework demonstrates the effectiveness of multi-view mutual distillation and model merging for RGB-Event representation learning, there remains room for further exploration. In this work, the distillation process mainly focuses on feature-level and logit-level knowledge exchange among the three event views, while more fine-grained distillation strategies, such as hierarchical feature alignment, uncertainty-aware peer teaching, and relation-aware cross-view transfer, are not fully investigated. Similarly, our model merging design adopts frequency-domain parameter decomposition and historical solution averaging, but more advanced merging mechanisms that explicitly model parameter conflicts, layer-wise reliability, or sample-adaptive merging coefficients could further improve the integration of multi-view knowledge.

\begin{figure}[!t]
\centering
\includegraphics[width=0.72\linewidth]{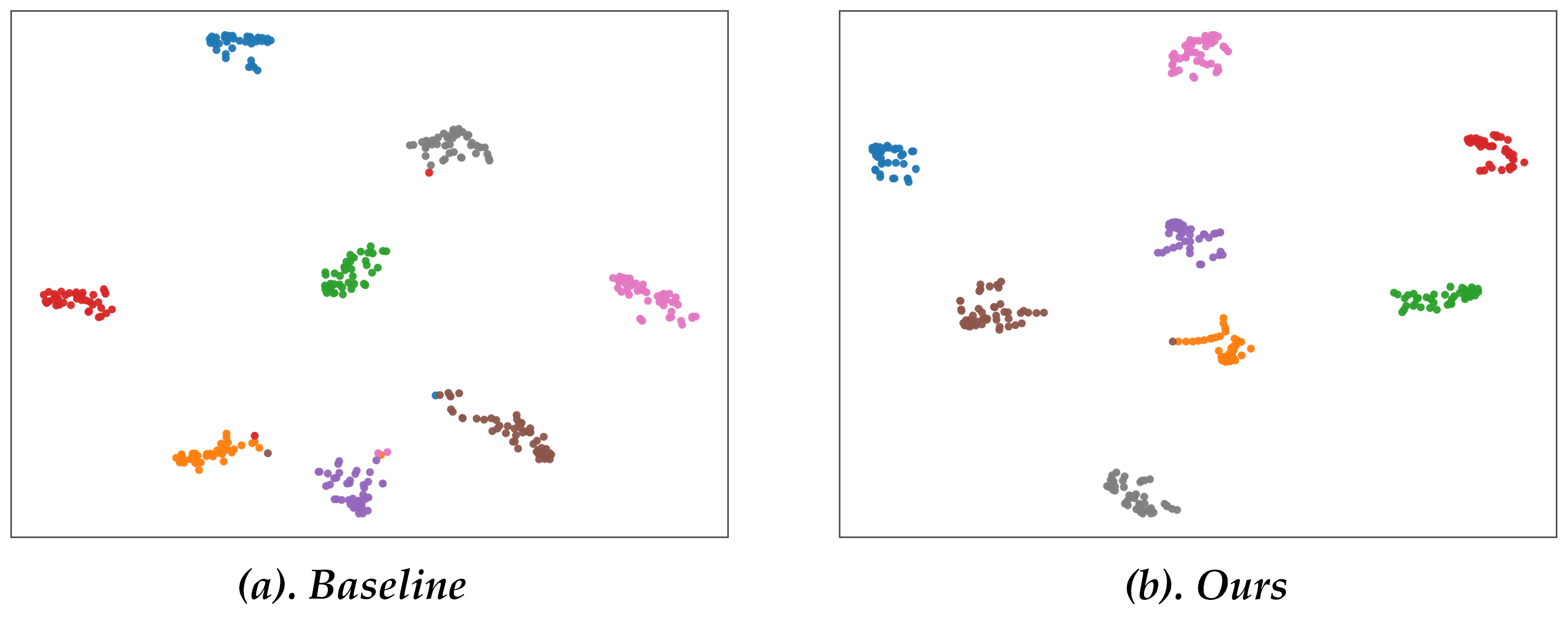}
\caption{The feature distribution of Baseline (left sub-figure) and our newly proposed model (right sub-figure) on the PokerEvent dataset using T-SNE.} 
\label{fig:tsne}
\end{figure}

\begin{figure}[!t]
\centering
\includegraphics[width=0.85\linewidth]{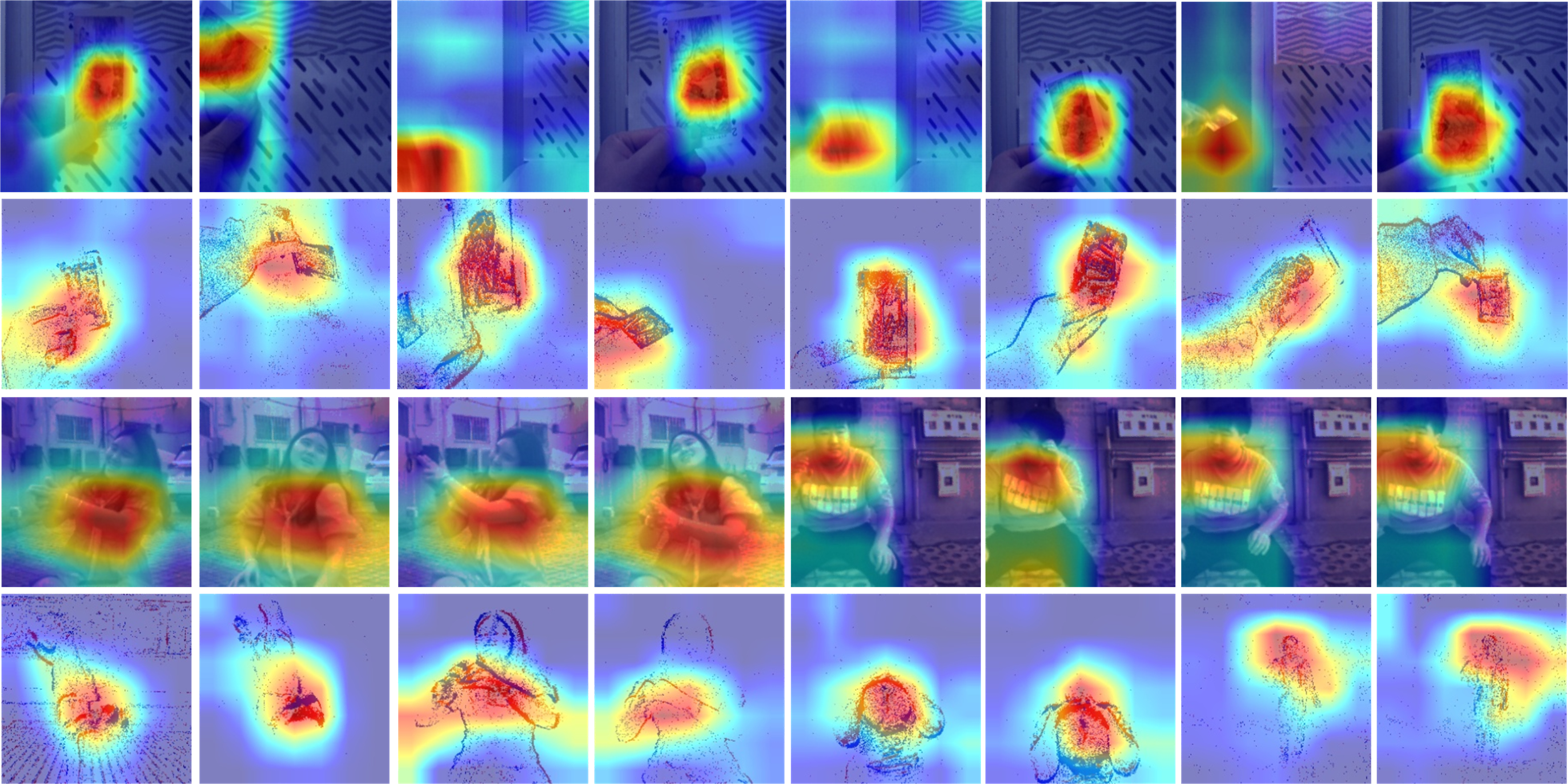}
\caption{Visualization of feature maps obtained using our proposed method on the PokerEvent and HARDVS dataset.} 
\label{fig:vis_map}
\end{figure}

\section{Visualization}
\noindent $\bullet$ \textbf{Top-5 Predictions on HARDVS and PokerEvent.} As shown
in Fig.~\ref{fig:Vis_Top-5}, we visualize the top-5 predictions for six representative samples from the HARDVS and PokerEvent datasets. The HARDVS samples correspond to action\_279, action\_291, and action\_294, while the PokerEvent samples correspond to Poker004, Koala, and Starfish. For all six samples, the ground truth category is ranked first. The results show that our method can reliably recognize human activities under challenging motion and illumination conditions and distinguish visually similar poker categories by effectively integrating complementary RGB and event information.

\noindent $\bullet$ \textbf{Distribution.} As shown in Fig.~\ref{fig:tsne}, we visualize the feature embeddings of the Baseline and our proposed method on the PokerEvent dataset, with different colors representing different test classes. Compared with the Baseline, which exhibits scattered feature distributions with weak intra-class compactness and limited inter-class separation, our method produces a more discriminative embedding space with compact class-wise clusters and clearer decision boundaries. These results indicate that the proposed module enhances feature discriminability, providing qualitative evidence for the performance improvements reported in our experiments.

\noindent $\bullet$ \textbf{Feature Maps.} As shown in Fig.~\ref{fig:vis_map}, we visualize the feature activation maps produced by the proposed method on representative samples from the PokerEvent and HARDVS datasets. For PokerEvent, the model consistently concentrates its responses on the card patterns and hand--object interaction regions, which contain the most discriminative cues for fine-grained category recognition. For HARDVS, the activated regions mainly cover the human body and motion-related areas, while irrelevant background responses are effectively suppressed. Moreover, the attention remains spatially consistent across different temporal samples and event representations. These qualitative results demonstrate that the proposed method can effectively integrate complementary appearance and motion information from RGB frames and heterogeneous event views, thereby learning more discriminative and robust representations for both fine-grained pattern recognition and human activity recognition.

\end{document}